\pdfoutput=1

\documentclass[11pt]{article}

\usepackage[final]{acl}

\usepackage{times}
\usepackage{latexsym}

\usepackage[T1]{fontenc}

\usepackage[utf8]{inputenc}

\usepackage{microtype}

\usepackage{inconsolata}

\usepackage{graphicx}

\usepackage{algorithm}
\usepackage{algorithmic}
\usepackage{multirow}
\usepackage{makecell}
\usepackage[most]{tcolorbox}
\usepackage{booktabs}
\usepackage{graphicx}
\usepackage{subcaption}
\usepackage{xcolor} 
\usepackage{tabularx}
\usepackage{multicol}
\usepackage{longtable} 
\usepackage{xcolor}
\usepackage{amsmath}  
\usepackage{tabularx}
\usepackage{adjustbox}
\usepackage{hyperref}

\newcommand{\green}[1]{\textcolor[rgb]{0,0.5,0}{#1}}

\definecolor{darkorange}{RGB}{255, 140, 0}
\definecolor{darkblue}{RGB}{84, 112, 198}
\definecolor{lightgreen}{RGB}{145, 204, 117}
\definecolor{lightyellow}{RGB}{250, 200, 88}
\definecolor{lightred}{RGB}{193, 78, 82}
\definecolor{lightblue}{RGB}{115, 192, 222}
\newtcolorbox{promptbox}[2][Prompt]{
colback=black!5!white,
arc=5pt, 
boxrule=0.5pt,
fonttitle=\bfseries,
title=#1, 
before upper={\small}, fontupper=\fontfamily{ptm}\selectfont,
colframe=#2,
}

\title{%
  \begin{tabular}{@{}c@{}l} 
    \raisebox{-0.3\totalheight}{\includegraphics[width=1.4cm]{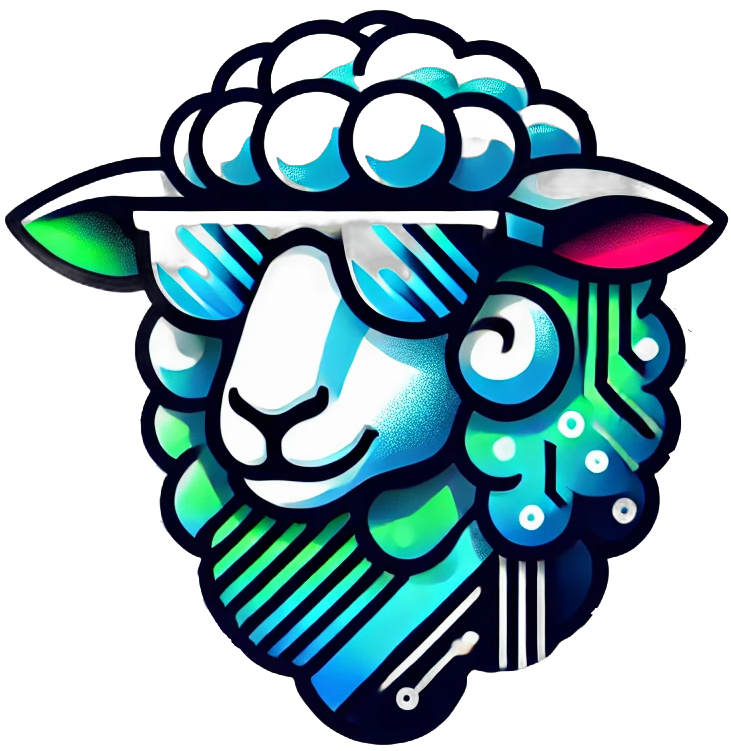}} &
    \begin{tabular}{@{}l@{}}
      \hspace{2.5mm}I-SHEEP: Self-Alignment of LLM from Scratch\\ through an Iterative Self-Enhancement Paradigm
    \end{tabular}
  \end{tabular}
}

\author{
    \textbf{Yiming Liang}\textsuperscript{1,2,8}\thanks{\ \ Equal contribution.}\quad
    \textbf{Ge Zhang}\textsuperscript{3,5,6}\footnotemark[1]\quad
    \textbf{Xingwei Qu}\textsuperscript{4,5,6}\footnotemark[1]\quad
    \textbf{Tianyu Zheng}\textsuperscript{5,6}\footnotemark[1] \\
    \textbf{Jiawei Guo}\textsuperscript{5,6}\quad
    \textbf{Xinrun Du}\textsuperscript{5,6}\quad
    \textbf{Zhenzhu Yang}\textsuperscript{5}\quad
    \textbf{Jiaheng Liu}\textsuperscript{5}\quad \\
    \textbf{Chenghua Lin}\textsuperscript{4}\quad 
    \textbf{Lei Ma}\textsuperscript{7,8}\quad
    \textbf{Wenhao Huang}\textsuperscript{6}\thanks{\ \ Corresponding authors.}\quad
    \textbf{Jiajun Zhang}\textsuperscript{1,2}\footnotemark[2]\\
     \textsuperscript{1}School of Artificial Intelligence, University of Chinese Academy of Sciences\\
     \textsuperscript{2}Institute of Automation, Chinese Academy of Sciences,
     \textsuperscript{3}University of Waterloo\\
     \textsuperscript{4}The University of Manchester,
     \textsuperscript{5}M-A-P,
     \textsuperscript{6}01.ai,
     \textsuperscript{7}Peking University,
     \textsuperscript{8}BAAI
\small \\
}

\begin{document}
\maketitle
\begin{abstract}
Large Language Models (LLMs) have achieved significant advancements, however, the common learning paradigm treats LLMs as passive information repositories, neglecting their potential for active learning and alignment. 
Some approaches train LLMs using their own generated synthetic data, exploring the possibility of active alignment.
However, there is still a huge gap between these one-time alignment methods and the continuous automatic alignment of humans.
In this paper, we introduce \textbf{I-SHEEP}, an \textbf{I}terative \textbf{S}elf-En\textbf{H}anc\textbf{E}m\textbf{E}nt \textbf{P}aradigm.
This human-like paradigm enables LLMs to \textbf{iteratively self-improve even in low-resource scenarios}.
Compared to the one-time alignment method Dromedary \cite{sun2023principledriven}, which refers to the first iteration in this paper, I-SHEEP can significantly enhance capacities on both Qwen and Llama models. 
I-SHEEP achieves a maximum relative improvement of 78.2\% in the Alpaca Eval, 24.0\% in the MT Bench, and an absolute increase of 8.88\% in the IFEval accuracy over subsequent iterations in Qwen-1.5 72B model.
Additionally, I-SHEEP surpasses the base model in various standard benchmark generation tasks, achieving an average improvement of 24.77\% in code generation tasks, 12.04\% in TrivialQA, and 20.29\% in SQuAD.
We also provide new insights based on the experiment results.
Our code, datasets, and models are available at \href{https://anonymous.4open.science/r/SHEEP/}{https://anonymous.4open.science/r/SHEEP/}.
\end{abstract}

\section{Introduction}
\begin{figure*}[!htb]
    \centering
\includegraphics[width=0.95\textwidth]{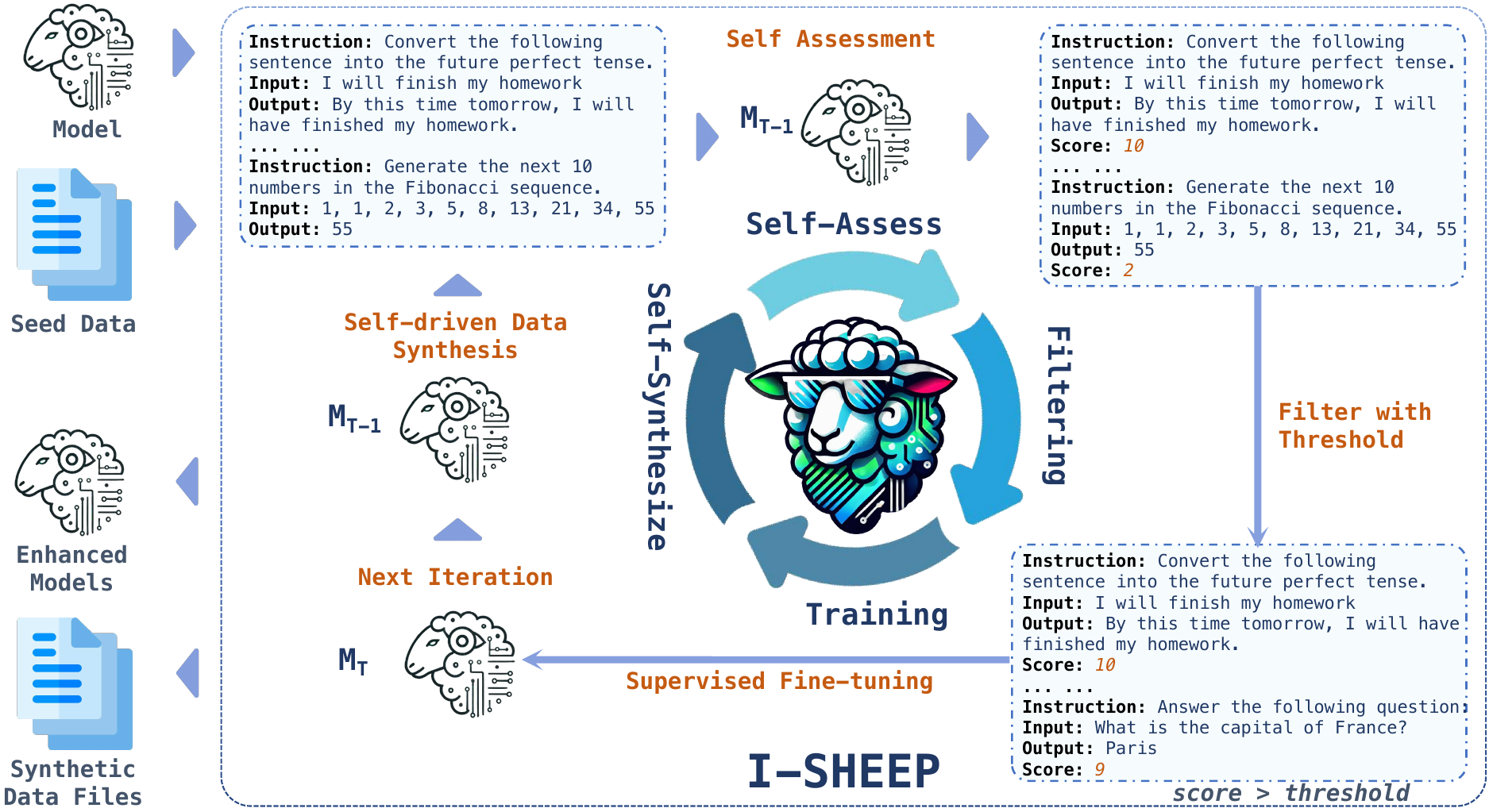}
  \caption{Pipeline of I-SHEEP. The I-SHEEP framework takes the base model and small seed dataset as input, aligns the base model iteratively from scratch independently, and finally obtains the self-enhanced models and high-quality synthetic datasets. The I-SHEEP framework consists of four main components: the self-synthesize process generates instruction-pair data, the self-assessment assesses the quality of the resulting data, the filtering component filters out low-quality data based on self-assessment, and the training component integrates the high-quality data into the base model.}
  \label{fig:I-SHEEP pipeline}
\end{figure*}

Early studies improve model performance using human-labeled data, but the high cost of labeling limits scalability \cite{zhou2024lima, zheng2024kun}. Some methods use powerful models to synthesize data, thereby improving student models \cite{alpaca, xu2024magpie}. However, these methods face performance ceilings and indirectly depend on strong models' reliance on human-labeled signals \cite{li2023selfalignment}. Additionally, they often treat models as passive information repositories, overlooking the models' ability to actively align. Other methods focus on the active alignment capabilities of LLMs, enhancing them through self-generated data. Nevertheless, these approaches typically rely on substantial external signals or tools, such as raw text \cite{li2023selfalignment}, retrieval-augmented generation (RAG) \cite{asai2023selfrag}, feedback from strong models \cite{lee2024llm2llm}, and high-quality questions \cite{huang2022large,wang2024steponfeet}, to achieve self-improvement.

Recently, some approaches explore the active alignment capabilities of LLMs in low-resource scenarios, aiming for models to self-improve with minimal reliance on external signals \cite{wang2022selfinstruct, sun2023principledriven, sun2023salmon}. For example, Self-Instruct \cite{wang2022selfinstruct} prompts the model to generate instructions using a seed dataset containing only 175 human-labeled instruction pairs, achieving self-alignment. Dromedary \cite{sun2023principledriven} uses 16 manually crafted principles to guide LLMs in generating instruction pair data, enhancing the quality of synthesized data. However, these methods are typically one-time alignment approaches, showing significant gaps compared to the continuous and automatic alignment that humans perform in varying environments. In this paper, we explore leveraging the model's internal metacognitive self-assessment to enable multi-round iterative self-improvement in low-resource settings, similar to human processes.

Educational research suggests that metacognitive self-assessment plays a vital role in continuous alignment, helping students reflect on their knowledge and skills, manage cognitive resources, and improve their performance \cite{Yan2023}. 
Inspired by this perspective, we introduce I-SHEEP, a human-like paradigm that enables LLMs to iteratively self-improve in low-resource settings.
As shown in Figure \ref{fig:I-SHEEP pipeline}, I-SHEEP begins with seed data and leverages the understanding and generation capabilities of LLMs to create additional instruction pairs. 
We then perform self-assessment, allowing LLMs to monitor and assess their learning process. 
By filtering out incorrect cognitions and retaining accurate ones, LLMs can self-improve by aligning themselves with these correct cognitions.
Through an iterative process, LLMs can continuously self-align, relying solely on their internal knowledge.

The main contributions can be summarized as follows: 
\textbf{(1) We introduce I-SHEEP, which aims to explore the potential of LLMs to iteratively self-improve in low-resource scenarios.} I-SHEEP incorporates metacognitive self-assessment to monitor and manage the learning process of LLMs, enabling iterative self-improvement. \textbf{(2) We analyze the factors that influence the continuous improvement potential of LLMs.} Our experiments show that the self-improvement ability of LLMs is influenced by their inherent capabilities and metacognitive levels, and varies with model size and metacognitive capacity. \textbf{(3) We validate the effectiveness and efficiency of the I-SHEEP framework through experiments.} Even in low-resource scenarios, I-SHEEP significantly improves the performance of LLMs on various chat benchmarks and standard benchmarks through multiple iterations.


\section{Related Work}
\subsection{Automatic Data Selection}
\citeauthor{zhou2024lima, bai2024coigcqia} emphasize that dataset quality outweighs quantity during the instruction fine-tuning stage. 
As a result, some studies on instruction data selection have emerged, focusing on identifying high-quality subsets from candidate datasets\cite{li2023quantity, du2023mods, liu2023makes, li2024superfiltering, ge2024clustering, xia2024less}.
These methods aim to improve the model performance, accelerate the training process, and facilitate data-efficient alignment.
\citeauthor{li2023quantity} introduce an Instruction-Following Difficulty (IFD) metric and use it to select the top 5\% of data for fine-tuning models.
The filtering phase in the I-SHEEP framework does not rely on predefined metrics, external models, or human assistance, and is orthogonal to existing selection methods.

\subsection{Synthetic Data for Improving Model}
Generating synthetic data refers to using the powerful generative capabilities of LLMs to create new data that simulates potential real-world scenarios, reducing the need for costly manual labeling. 
Some methods use the model's self-generated data to improve itself \cite{wang2022selfinstruct, sun2023principledriven, sun2023salmon, yehudai2024genie}. 
Other methods leverage powerful closed models to generate synthetic data, enhancing the capabilities of open-source models \cite{alpaca, vicuna2023, xu2023wizardlm, yu2023wavecoder, wei2023magicoder}. 
In addition to generating complete instruction-output pairs, some methods collect existing raw data and synthesize corresponding questions or answers to create supervised data for improving the model \cite{huang2022large, li2023selfalignment, zheng2024kun, mitra2024orcamath, wang2022selfconsistency, asai2023selfrag}.
Some methods begin with instruction-output pairs, generating feedback or refining answers to improve data quality and enhance the model's reasoning capabilities.\cite{lu2023self, li2024humans, gou2023tora}. 
The I-SHEEP framework evolves from the aforementioned static, one-time improvement paradigm to a dynamic, continuous self-enhancement process.

\subsection{Iterative Enhancement for LLMs}
There are several approaches to iterative enhancement that rely on the help of strong models or external tools \cite{chen2024iteralign, chen2023iterative, lu2023self, gao2023confucius, lee2024llm2llm}.
IterAlign \cite{chen2024iteralign} employs strong models like GPT-4 and Claude2 to detect and correct errors in responses from base LLMs and give the corresponding constitution for improving the safety of LLMs.
These methods in iterative enhancement typically depend on strong models or external tools to guarantee ongoing model optimization and avoid model collapse. 
In addition, some methods explore iterative enhancement in the RLHF phase to continuously align the model with human preference \cite{yuan2024selfrewarding, liu2024iterative, pang2024iterative, xu2024dpo, xu2023things, wu2024metarewarding, wang2024steponfeet}. 
These iterative RLHF methods start with the aligned model, while we focus on the base model continuous self-alignment from scratch.

\section{Methodology}

\subsection{Self-Driven Data Synthesis}

Self Instruct \cite{wang2022selfinstruct} leverages an off-the-shelf large language model (LLM) for the generation of synthetic data. 
The approach starts with a small set of 175 prompts, known as the seed task pool, leveraging the model's powerful understanding and generative capabilities to generate a broader range of prompts and responses.
This section elaborates on the Self-Driven Data Synthesis process from two perspectives: Instruction generation and response generation. 
For ease and consistency in data creation, we utilize a standardized instruction format introduced by Alpaca \cite{alpaca}, enabling the direct generation of instructions along with their corresponding potential inputs. 

\textbf{Instruction generation.} Having some prompts from the seed dataset $D^s$ and the meta-prompt $p^{meta}$ from Alpaca \cite{alpaca}. 
The process that model $M$ generating new prompt set $\mathcal{P}$ through In-Context Learning (ICL) can be modeled as: 
$$p_i = argmax_p(p_i|\{d\}, p^{meta}; \theta)$$
$p_i$ denotes a new prompt generated by model $M$, $\{d\}$ represents a subset sampled from the seed dataset $D^s$ for in-context learning (ICL). 
The symbol $\theta$ stands for the parameter of model $M$.

\textbf{Response generation. }After obtaining the set of prompts $\mathcal{P}$, we use the model $M$ to generate corresponding responses $\mathcal{R}$ via a zero-shot approach. 

\subsection{Self-Assessment and Data Filtering}

To ensure that the data used for self-enhancement maintains a high-quality standard, a two-stage process comprising self-assessment and data filtering is implemented.

\textbf{Self-Assessment. }We pair the generated prompt set $\mathcal{P}$ and response set $\mathcal{R}$ to form the instruction-output pair data $D_{raw}$.
Given the capacity limitations of models, ensuring the quality of synthetic pairs can be challenging, making it essential to assess the quality of the generated data. 
Manual assessment is often impractical, therefore, we introduce an automated assessment method that relies solely on the model.
Specifically, the model autonomously evaluates each generated response for its quality and adherence to the instructions. 
Each entry is scored based on predefined criteria, which quantitatively reflect the compliance and quality of the response.


\textbf{Data Filtering. }
After the self-assessment, the subsequent data filtering phase discards entries that do not meet the specified quality threshold.
This step guarantees that only entries of the highest quality are retained in the dataset, thereby enhancing the overall reliability and utility of the generated data.
Initially, we apply heuristic rule-based filtering to the generated data during data generation, following the Self-Instruct \cite{wang2022selfinstruct}.
Additionally, after data generation, we filter the instruction-output pairs based on the assessment scores from the self-assessment phrase.
A threshold $\mathcal{C}$ is applied to filter $D_{raw}$ based on assessment scores, yielding a high-quality dataset $D$. 


\subsection{Iterative Model Enhancements}

The Iterative Self-Enhancement algorithm aims to incrementally enhance a language model by generating and utilizing high-quality synthetic datasets. 
As shown in Algorithm \ref{alg:SHEEP}, starting with an initial model $M^{base}$ and a small seed task set $D^s$, the algorithm iterates over a specified number of steps $\mathcal{T}$ and a filtering threshold $\mathcal{C}$. 
At each iteration $t$, the algorithm performs several functions: it generates a new set of prompts, $\mathcal{P}^t$, using a prompt generation process that leverages the current model $M^t$ and the seed data $D^s$. 
It then produces corresponding responses, $\mathcal{R}^t$, forming a raw dataset, $D^t_{raw}$ = $\{\mathcal{P}^t, \mathcal{R}^t\}$. 
This dataset undergoes a self-assessment process to evaluate the quality of responses, after which it is filtered using the threshold $\mathcal{C}$ to retain only high-quality data, resulting in $D^t$. 
The model $M^t$ is then trained on $D^t$ to align it closely with the refined data, enhancing its performance iteratively by supervised fine-tuning (SFT) approach. 
This process continues until it concludes at step $\mathcal{T}$, ultimately producing a stronger language model $M^\mathcal{T}$ and a refined synthetic dataset $D^\mathcal{T}$.

\section{Experiments}
\subsection{Evaluation}

\subsubsection{Chat Evaluation}
We evaluate the instruction-following ability and response quality of aligned models with three chat benchmarks, AlpacaEval\cite{dubois2023alpacafarm}, MT-Bench\cite{zheng2024judging}, and IFEval\cite{zhou2023instructionfollowing}, due to their comprehensiveness, fine granularity, and reproducibility.
Both AlpacaEval and MT-Bench rely on GPT as an evaluator.
IFEval provides four types of accuracy scores: prompt-level strict-accuracy, inst-level strict-accuracy, prompt-level loose-accuracy, and inst-level loose-accuracy.

\begin{algorithm}[!tb]
\caption{Iterative Self-Enhancement Algorithm}
\label{alg:SHEEP}
\textbf{Input}: Initial seed task set $D^s$, Base model $M^{base}$\\
\textbf{Hyper-parameter}: Iteration steps $\mathcal{T}$, Filtering threshold $\mathcal{C}$, Data size $\mathcal{I}$\\
\textbf{Output}: Enhanced LLMs $M^{\mathcal{T}}$, High-quality datasets $D^{\mathcal{T}}$
\begin{algorithmic}[1]
\STATE Initialize $M^0 \leftarrow M^{base}$
\FOR{$t = 0$ \TO $\mathcal{T}$}
    \STATE $\mathcal{P}^t \leftarrow$ generate\_prompts($D^s$, $p^{meta}$, $M^t$)
    \STATE $\mathcal{R}^t \leftarrow$ generate\_responses($\mathcal{P}^t$, $M^t$)
    \STATE $D^t_{raw} \leftarrow \{(\mathcal{P}^t, \mathcal{R}^t)\}$
    \STATE $S^t \leftarrow$ self\_assessment($D^t_{raw}$, $M^t$)
    \STATE $D^t \leftarrow$ filtering($D^t_{raw}$, $S^t$, $\mathcal{C}$)
    \STATE $M^{t+1} \leftarrow$ SFT($M^{base}$, $D^t$)
\ENDFOR
\RETURN $M^t, D^t$
\end{algorithmic}
\end{algorithm}

\subsubsection{OpenCompass Evaluation}
We use the OpenCompass evaluation platform \cite{2023opencompass}, a comprehensive one-stop platform for LLM evaluation. The evaluation includes standard benchmarks such as BoolQ \cite{clark2019boolq}, PIQA \cite{bisk2019piqa}, SIQA \cite{sap2019socialiqa}, HellaSwag \cite{zellers2019hellaswag}, WinoGrande \cite{sakaguchi2019winogrande}, ARC-c \cite{clark2018think}, OpenBookQA-Fact \cite{mihaylov-etal-2018-suit}, CommonsenseQA \cite{2023opencompass}, and MMLU \cite{hendrycks2020measuring}. 
It also includes code generation benchmarks such as HumanEval \cite{chen2021evaluating} and MBPP \cite{austin2021program}, word knowledge benchmark TriviaQA \cite{joshi2017triviaqa}, and reading comprehension benchmark SQuAD2.0 \cite{rajpurkar2018know}.
Full results on these benchmarks are available in Appendix \ref{appendix:opencompass_more_results}.

\subsection{Main Settings} 
We conduct experiments on the Qwen-1.5 \cite{qwen1.5} and Llama-3 \cite{dubey2024llama} models to validate the effectiveness and generalization of I-SHEEP. 
Additionally, we explore the impact of different model sizes on I-SHEEP by conducting experiments on Qwen-1.5 1.8B, 4B, 7B, 14B, 32B, and 72B models, providing a detailed analysis based on the experimental results.
In each iteration, the dataset for training is generated by the model from the last iteration. The case study of the generated data and the overall quality analysis can be found in Appendix \ref{appendix:case study} and Appendix  \ref{appendix:data_quality_analysis}, respectively.
We utilized LLaMA-Factory \cite{zheng2024llamafactory} for LoRA fine-tuning, with specific parameters detailed in Appendix  \ref{appendix:hyperparameters}. 
Under the configuration of using VLLM for inference \cite{kwon2023efficient}, the maximum duration of each iteration is about 4 hours on NVIDIA A800-SXM4-80GB$\times$8, equivalent to one iteration time for Qwen-1.5 72B.

\subsection{Self-Assessment and Filter Settings}
During the self-assessment phase, we propose three variants, simple standard prompt, combined standard prompt, and ICL prompt, to evaluate data quality. 
Detailed prompt contents can be found in Appendix \ref{appendix: prompt_content}.

In the filtering phase, there are six settings, simple standard prompt based filtering, combined standard prompt based filtering, ICL prompt filtering, PerPLexity (PPL) filtering, density filtering, and the combination of density and PPL filtering.
In addition to the first three filtering settings based on scores obtained in the Self-Assessment phase, we also explore data filtering methods that do not rely on external tools or models.
For example, PPL filtering uses the PPL value computed by the model itself to evaluate the quality of instruction-output pairs, thereby eliminating low-quality data.
We filter out data points with PPL greater than 50.
Density filtering extracts vector representations from the model's final layer and performs K Nearest Neighbors (KNN) clustering, sampling from each cluster to ensure dataset diversity.
We set 3000 as the clustering number K.
The combination of density and PPL filtering setting first clusters the data and then selects samples with lower PPL values from each cluster, ensuring the filtered dataset's quality and diversity.

\subsection{Baseline}
We use the base model, Self Instruct \cite{wang2022selfinstruct}, and Dromedary \cite{sun2023principledriven} as baselines to explore the continuous and automatic enhancement of the human-like framework, I-SHEEP. 
Self Instruct is a one-time alignment approach where LLMs are trained directly on data they generate, without a self-assessment phase.
Similarly, Dromedary is a one-time alignment process where the model generates responses following specific principles, which are then engraved into the model. 
This approach is similar to the first iteration setting described in this paper.

\subsection{Iterative Settings and Ablation Settings}
\textbf{Iterative Settings. }We investigate the impact of I-SHEEP on efficiency across different iterative self-enhancement settings, including using data generated by the last iteration model to train the base model, using data generated by the last iteration model to train the last iteration model, and using data generated by all previous iterations to train the base model.
Additionally, we directly generate 20K and 30K data points for comparative experiments to eliminate the influence of data size in the iterative settings mentioned above.
Notably, in the first iteration, all settings are identical, where the base model generates 10k data, filters it, and uses it to fine-tune itself, akin to the Dromedary\cite{sun2023principledriven}.

\begin{table*}[!htb]
\centering
\resizebox{\textwidth}{!}{
    \begin{tabular}{@{}lccccccccccc@{}}
    \toprule
        \multicolumn{2}{c}{\textbf{Setting}} &\multicolumn{6}{c}{\textbf{Chat Benchmark}}&\multicolumn{4}{c}{\textbf{Standard Benchmark}}\\ \cmidrule(r){1-2} \cmidrule(r){3-8} \cmidrule(r){9-12}
        & & \multirow{3}{*}{\makecell{Alpaca\\Eval}} & \multirow{3}{*}{\makecell{MT\\Bench}} & \multicolumn{4}{c}{IFEval} & \multicolumn{2}{c}{Code} & Knowledge & Reading Comprehension\\ \cmidrule(r){5-8} \cmidrule(r){9-10} \cmidrule(r){11-11} \cmidrule(r){12-12}
        & & & & \makecell{P-level\\S-accuracy} & \makecell{ I-level\\S-accuracy} & \makecell{P-level\\L-accuracy} & \makecell{ I-level\\l-accuracy} & \makecell{Human\\Eval/Plus} & MBPP & \makecell{Trivia\\QA} & SQuAD 2.0 \\
    \midrule
        \multirow{4}{*}{\textbf{1.8B}}
        & base & -- & -- & -- & -- & -- & -- & 6.71/6.10 & 16.40 & \textbf{31.18} & \textbf{30.02} \\
        & iter1 & 1.51 & \textbf{3.76} & 15.53 & 25.30 & 17.74 & 28.06 & 11.59/9.15 & 16.80 & 19.38 & 13.16 \\
        & \textbf{iter2} & 1.54 & 3.53 & \textbf{16.27} & \textbf{27.10} & \textbf{19.22} & \textbf{31.41} & \textbf{15.24/12.20} & 17.40 & 16.88 & 14.57 \\
        & iter3 & \textbf{2.30} & 3.16 & 13.68 & 24.46 & 15.34 & 27.22 & 14.02/10.98 & \textbf{17.80} & 12.49 & 13.91 \\
    \midrule
        \multirow{4}{*}{\textbf{4B}}
        & base & -- & -- & -- & -- & -- & -- & 10.98/8.54 & 28.00 & \textbf{40.95} & 27.96\\
        & iter1 & 2.61 & 4.97 & 19.41 & 29.98 & 24.03 & 34.77 & 30.49/26.83 & 34.00  & 38.94 & 24.90 \\
        & \textbf{iter2} & 2.96 & 4.79 & \textbf{19.78} & \textbf{32.61} & \textbf{23.84} & \textbf{36.81} & 31.10/27.44 & 35.20 & 37.20 & 24.63 \\
        & \textbf{iter3} & \textbf{3.78} & \textbf{4.99} & 18.85 & 31.41 & 22.18 & 35.37 & \textbf{32.93/28.66} & \textbf{35.80} & 35.37 & \textbf{31.67} \\
    \midrule
        \multirow{4}{*}{\textbf{7B}}
        & base & -- & -- & -- & -- & -- & -- & 10.98/8.54 & 36.60 & \textbf{51.00} & 33.14\\
        & iter1 & 5.19 & 5.08 & 28.47 & 39.93 & 31.05 & 43.41 & 45.73/39.63 & \textbf{41.20} & 45.81 & 26.36 \\
        & \textbf{iter2} & \textbf{5.37} & \textbf{5.13} & \textbf{30.13} & \textbf{40.89} & \textbf{33.09} & \textbf{43.88} & \textbf{47.56/42.68} & 41.00 & 42.83 & 28.36\\
        & iter3 & 5.22 & 4.97 & 29.21 & 40.29 & 30.68 & 43.05 & 45.12/40.24 & 40.60 & 40.53 & \textbf{33.76}\\
    \midrule
        \multirow{4}{*}{\textbf{14B}}
        & base & -- & -- & -- & -- & -- & -- & 17.68/15.85 & 41.40 & \textbf{57.72} & 20.37  \\
        & iter1 & 4.77 & 5.68 & 28.84 & 41.13 & \textbf{33.46} & \textbf{46.40} & 45.73/40.85 & \textbf{49.00} & 56.81 & 30.52 \\
        & \textbf{iter2} & 6.27 & \textbf{5.97} & \textbf{30.87} & 42.93 & \textbf{33.46} & \textbf{46.40} & \textbf{48.78/42.07} & 45.60 & 54.45 & 38.57\\
        & \textbf{iter3} & \textbf{7.30} & 5.48 & 30.13 & \textbf{43.05} & 33.27 & 46.04 & \textbf{50.00/43.29} & 45.20 & 55.30 & \textbf{43.42}\\
    \midrule
        \multirow{5}{*}{\textbf{32B}}
        & base & -- & -- & -- & -- & -- & -- & 22.56/21.34 & \textbf{47.40} & 65.88 & 29.56\\
        & iter1 & 8.27 & 5.56 & 33.46 & 45.32 & 37.52 & 50.12 & \textbf{58.54/51.83} & 44.20 & \textbf{60.81} & 41.34\\
        & iter2 & 8.26 & 5.68 & 36.04 & 47.60 & \textbf{39.56} & \textbf{51.92} & 56.71/50.61 & 41.80 & 59.43 & 42.15\\
        & \textbf{iter3} & \textbf{9.30} & \textbf{5.69} & \textbf{36.41} & \textbf{47.96} & 38.82 & 51.56 & 56.71/51.83 & 42.20 & 59.73 & 44.04\\
        & iter4 & 8.64 & 5.62 & 33.83 & 46.88 & 38.45 & 51.56 & 56.10/50.61 & 40.60 & 58.95 & \textbf{47.07}\\
    \midrule
        \multirow{6}{*}{\textbf{72B}} 
        & \textcolor{red}{iter1} & 6.64\hspace{2pt}\green{$\uparrow$5.19} & 6.43\hspace{2pt}\green{$\uparrow$1.54} & 35.67\hspace{2pt}\green{$\uparrow$8.88} & 49.16\hspace{2pt}\green{$\uparrow$6.72} & 40.48\hspace{2pt}\green{$\uparrow$7.02} & 53.96\hspace{2pt}\green{$\uparrow$4.79} & 50.61/45.12\hspace{2pt}\green{$\uparrow$6.10/8.54}  & 51.20\hspace{2pt}\green{$\uparrow$4.80} & 60.81\hspace{2pt}\green{$\uparrow$9.62} & 50.68\hspace{2pt}\green{$\uparrow$17.27}\\
        & iter2 & 9.06 & 7.90 & 37.34 & 51.32 & 40.85 & 54.56 & 56.71/49.39 & 51.80 & 61.55 & 52.27\\
        & iter3 & 10.51 & \textbf{7.97} & 41.22 & 54.32 & 44.18 & 57.19 & 56.10/50.61 & 52.60 & 62.00 & 61.42\\
        & iter4 & 11.22 & 5.45 & 42.14 & 54.56 & 46.21 & 58.63 & 51.83/47.56 & \textbf{56.00} & \textbf{70.43} & 64.55\\ 
        & \textbf{iter5} & \textbf{11.83} & 5.62 & \textbf{44.55} & \textbf{55.88} & \textbf{47.50} & \textbf{58.75} & \textbf{56.71/53.66} & 55.60 & 70.11 & \textbf{67.95}\\
        & iter6 & 11.60 & 5.75 & 42.33 & 53.84 & 45.10 & 56.95 & 51.22/48.17 & 55.20 & 70.01 & 67.82\\
    \midrule   
    \multicolumn{2}{l}{\textcolor{red}{Base Model}} & -- & -- & -- & -- & -- & -- & 21.34/20.12\hspace{2pt}\green{$\uparrow$35.37/33.54} & 50.20\hspace{2pt}\green{$\uparrow$5.80} & 58.07\hspace{2pt}\green{$\uparrow$12.36} & 47.66\hspace{2pt}\green{$\uparrow$20.29}\\
    
    \multicolumn{2}{l}{\textcolor{red}{Self Instruct}} & 5.26\hspace{2pt}\green{$\uparrow$6.57} & 7.82\hspace{2pt}\green{$\uparrow$0.15} & 33.64\hspace{2pt}\green{$\uparrow$10.91} & 47.60\hspace{2pt}\green{$\uparrow$8.28} & 39.56\hspace{2pt}\green{$\uparrow$7.94} & 53.00\hspace{2pt}\green{$\uparrow$5.75} & 53.05/46.95\hspace{2pt}\green{$\uparrow$3.66/6.71} & 48.40\hspace{2pt}\green{$\uparrow$7.60} & 71.25\hspace{2pt}\green{$\downarrow$-0.82} & 51.90\hspace{2pt}\green{$\uparrow$16.05}\\
    \bottomrule
    \end{tabular}
}
\caption{Main results: experimental performance of various model sizes across different iteration steps. We stop the iteration when the performance improvement in subsequent iterations stagnates or diminishes. The red settings represent the baseline for our experiments on Qwen-1.5 72B. The \textcolor{red}{Self Instruct} \cite{wang2022selfinstruct} setting involves training the model using generated data without filtering. The \textcolor{red}{iter1} setting indicates training the model using filtered data, which is selected based on prompts, similar to the Dromedary approach \cite{sun2023principledriven}. \textbf{Bold} results indicate the best results and \green{$\uparrow$green} values represent the maximal improvement over the baseline in subsequent iterations.}
\label{table:main_results}
\end{table*}

\begin{figure*}[!htb]
    \centering
    \begin{subfigure}[b]{0.49\textwidth}
        \centering
        \includegraphics[width=\textwidth]{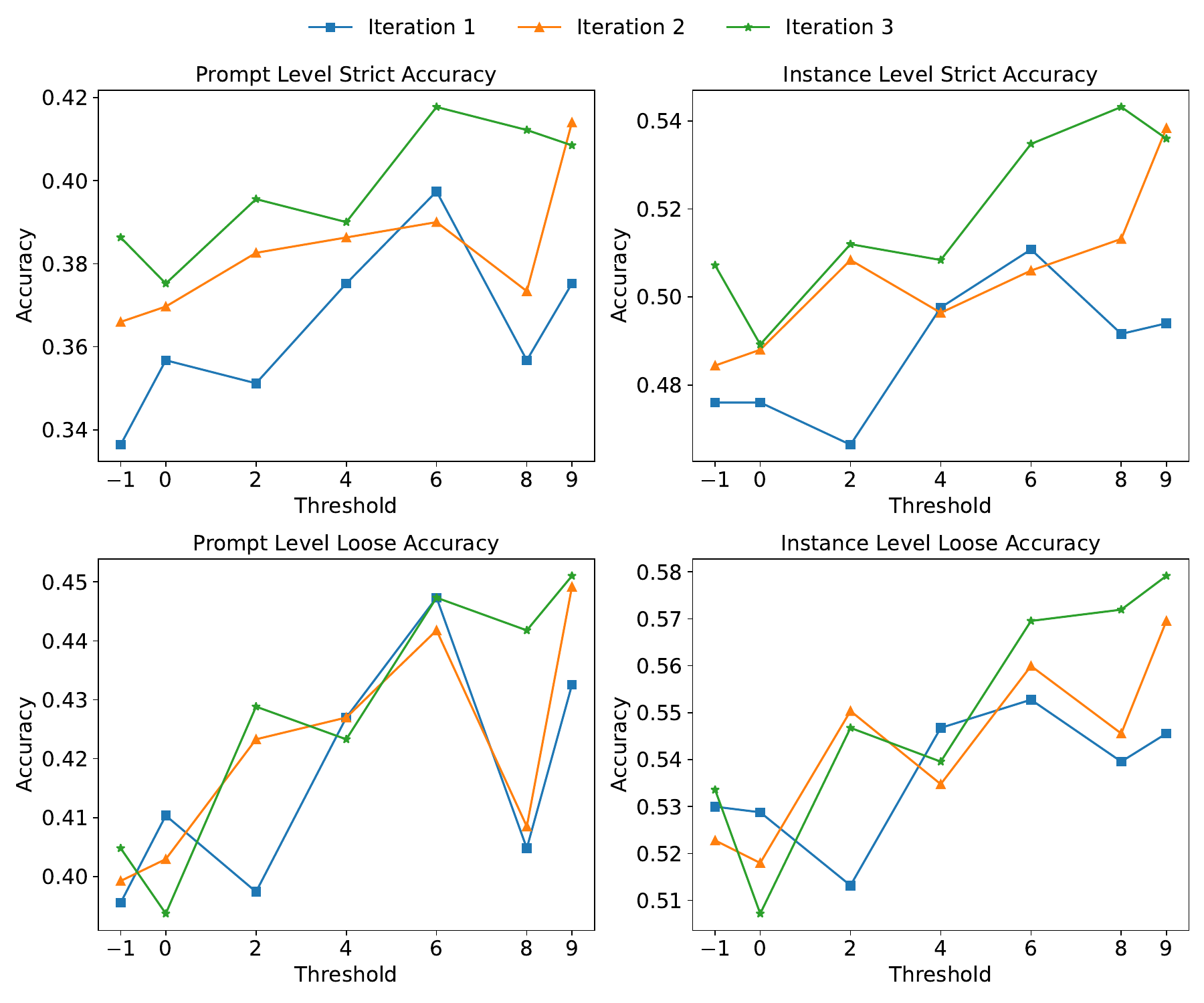}
        \caption{Performance in the first three iterations with different thresholds. }
        \label{fig:threshold_total}
    \end{subfigure}
    \hfill
    \begin{subfigure}[b]{0.49\textwidth}
        \centering
        \includegraphics[width=\textwidth]{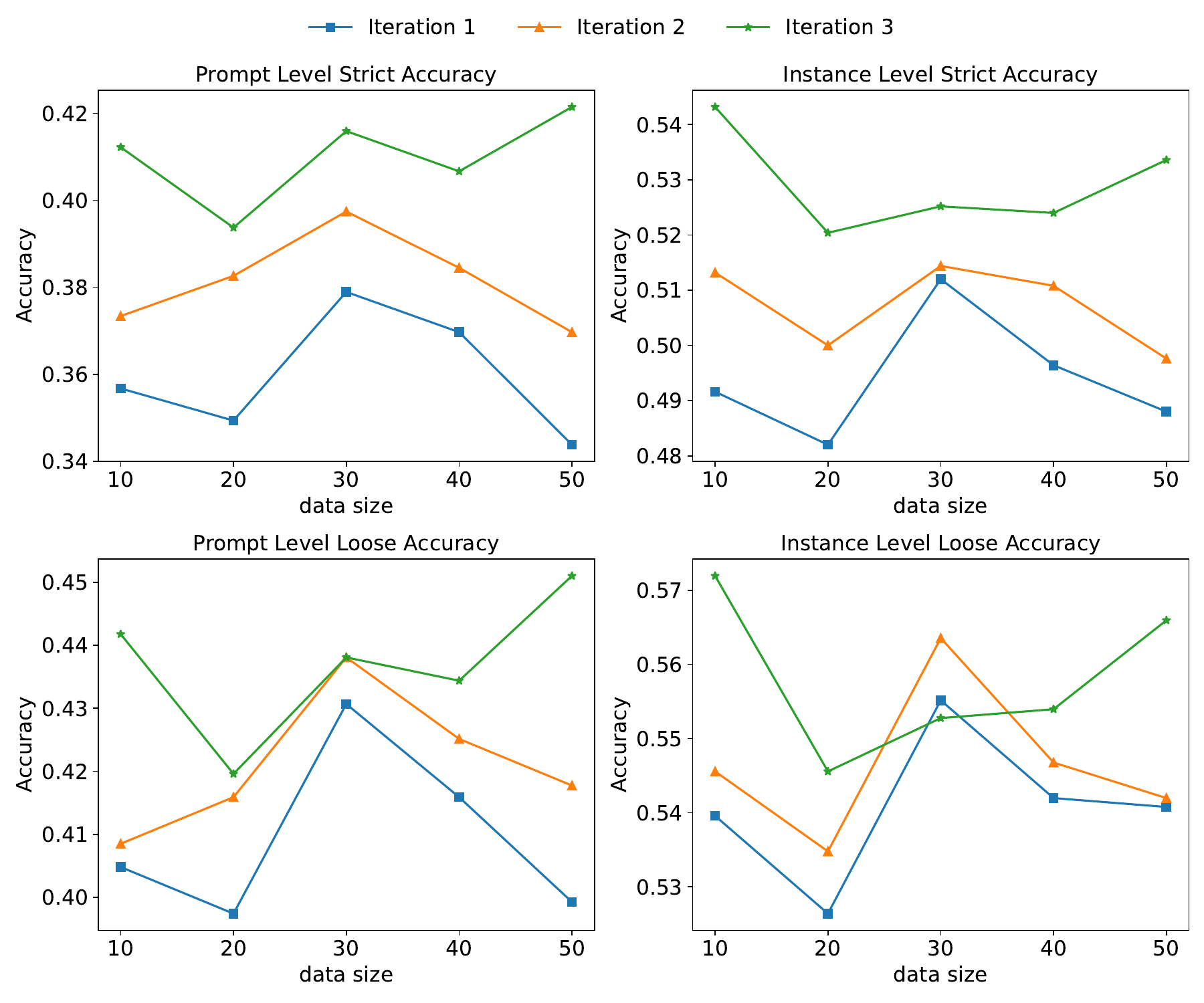}
        \caption{Performance in the first three iterations with different data sizes.}
        \label{fig:data_size_total}
    \end{subfigure}
    \caption{Ablation performance for the first three iterations across different thresholds and data sizes. In subfigure \ref{fig:threshold_total}, the threshold -1 means that the generated data is not filtered by heuristic rules. The threshold 0 represents that the I-SHEEP process does not use the self-assessment phase. Other thresholds represent filtering low-quality data using the threshold, which refers to the score from the self-assessment phase. In subfigure \ref{fig:data_size_total}, the values on the horizontal axis represent the amount of data generated (in thousands).}
    \label{fig:combined_performance}
\end{figure*}

\textbf{Ablation Settings. }we adjust high-dimensional variables such as the threshold $\mathcal{C}$ in the self-assessment phase, data size $\mathcal{I}$ in the generation phase, and iteration steps $\mathcal{T}$ in the iterative training phase to validate their impact on I-SHEEP.
Furthermore, we conduct ablation experiments with different levels of metacognitive self-assessment, including no self-assessment, assessing only response quality, assessing only instruction-following degree, and assessing both response quality and instruction-following degree.

\section{Results}

\subsection{Main Results}
Table \ref{table:main_results} shows the experimental performance of various model sizes across different iteration steps. 
There are some new findings:
\textbf{(1) I-SHEEP exhibits efficacy across various model sizes, with particularly notable improvements in 72B.}
I-SHEEP achieves a maximum relative improvement of 78.2\% in the Alpaca Eval, 24.0\% in the MT Bench, and an absolute increase of 8.88\% in the IFEval prompt-level strict accuracy over subsequent iterations in Qwen-1.5 72B model.
Additionally, I-SHEEP surpasses the base model in various standard benchmark generation tasks, achieving an average improvement of 24.77\% in code generation tasks, 12.04\% in Trivial QA, and 20.29\% in SQuAD.
we find that the scores for the second round of dialogues drop significantly after the fourth iteration. 
This decline is likely due to our generated data consisting solely of single-round dialogues, which do not improve and may even harm the scores for the second round of dialogues.
More analysis can be found in the Appendix \ref{appendix:MT_Bench}.
\textbf{(2) The potential for improvement varies with different model sizes.}
The 1.8B, 4B, 7B, and 14B models exhibit improvements over two iterations, 32B and 72B model can improve three and five iterations, respectively, according to the IFEval benchmark.

\subsection{Iterative Setting Results}

\begin{table}[!htb]
\centering
\resizebox{\columnwidth}{!}{
    \begin{tabular}{@{}cccccccc@{}}
    \toprule
        \multicolumn{2}{c}{\textbf{Setting}} & \multicolumn{6}{c}{\textbf{Chat Benchmark}} \\ \cmidrule(r){1-2} \cmidrule(r){3-8}
        & & \multirow{3}{*}{\makecell{Alpaca\\Eval}} & \multirow{3}{*}{\makecell{MT\\Bench}} & \multicolumn{4}{c}{IFEval} \\ \cmidrule(r){5-8}
        & & & & \makecell{P-level\\S-accuracy} & \makecell{ I-level\\S-accuracy} & \makecell{P-level\\L-accuracy} & \makecell{ I-level\\L-accuracy} \\ 
    \midrule
        \multicolumn{2}{c}{iter1(Dromedary)} & 6.64 & 6.43 & 35.67 & 49.16 & 40.48 & 53.96 \\
    \midrule
        \multirow{2}{*}{Direct} 
        & 20k & 7.18 & 7.87 & 39.37 & 50.72 & 43.25 & 54.56 \\
        & 30k & 6.53 & 7.75 & 38.08 & 50.24 & 43.07 & 54.92 \\
    \midrule
        \multirow{2}{*}{\makecell{Total\_base}}
        & iter2 & 7.25 & 7.94 & 39.00 & 50.72 & 45.47 & 56.47 \\
        & iter3 & 7.51 & 7.94 & 37.52 & 48.32 & 41.59 & 52.76 \\
    \midrule
        \multirow{2}{*}{\makecell{One\_last}}
        & iter2 & 7.76 & 7.76 & 38.45 & 50.48 & 41.96 & 54.92 \\
        & iter3 & 8.45 & 7.82 & 38.63 & 51.80 & 42.70 & 56.12 \\
    \midrule
        \multirow{2}{*}{\textbf{One\_base}}
        & iter2 & 9.06 & 7.90 & 37.34 & 51.32 & 40.85 & 54.56 \\
        & \textbf{iter3} & \textbf{10.51} & \textbf{7.97} & \textbf{41.22} & \textbf{54.32} & \textbf{44.18} & \textbf{57.19} \\
    \bottomrule
    \end{tabular}
}
\caption{The performance of various iteration settings at different iteration steps. \textit{One\_base} and \textit{One\_last} means using data from the last iteration to train the base and the last iteration model respectively. \textit{Total\_base} means using data from all previous iterations to train the base model. \textit{Direct} represents using data generated by the base model to train itself.}
\label{table:iterative_setting}
\end{table}

Table \ref{table:iterative_setting} presents the chat benchmark performance for the Qwen-1.5 72B model across various iteration settings. 
More benchmark results are available in Appendix \ref{appendix:opencompass_more_results}.
Our findings are as follows:
\textbf{(1) Training the base model with data from the last iteration model is effective for iterative self-enhancement.} 
At the third iteration in the One\_base Setting, training the base model with the last iteration data achieves the highest performance on the chat benchmark.
The notable performance improvement under this setting suggests that the model has the potential for further enhancement (refer to Table \ref{table:main_results} 72B results). Therefore, we chose the \textit{One\_base} setting for all subsequent experiments.
\textbf{(2) The data size is not the main factor influencing iterative improvement. }
Training the base model with the last iteration data at the 3rd iteration outperforms training the base model with a combination of all data from previous iterations. 

\subsection{Threshold Ablation}
As shown in Figure \ref{fig:threshold_total}, as the threshold increases, the performance of I-SHEEP at the 3rd iteration shows an upward trend.
The threshold 8 is selected to ensure the possibility of further iterative improvement, given the significant performance increase in iteration 2 and iteration 3, and the good performance at iteration 3 with a threshold of 8.
Choosing a threshold of 8 is not necessarily the optimal experimental setting, as thresholds of 6, 7, 8, and 9 are all possible. 


\begin{table}[!htb]
\centering
\resizebox{\columnwidth}{!}{
    \begin{tabular}{@{}cccccc@{}}
    \toprule
        \multicolumn{2}{c}{Setting}&\multicolumn{4}{c}{\textbf{Chat Benchmark}}\\ \cmidrule(r){1-2} \cmidrule(r){3-6}
        & & \multicolumn{4}{c}{IFEval} \\ \cmidrule(r){3-6}
        & & \makecell{P-level\\S-accuracy} & \makecell{ I-level\\S-accuracy} & \makecell{P-level\\L-accuracy} & \makecell{ I-level\\L-accuracy} \\ 
    \midrule
        \multirow{3}{*}{Density} 
        & iter1 & 34.20 & 46.76 & 39.56 & 51.80 \\
        & iter2 & 37.34 & 49.76 & 41.22 & 53.72 \\
        & iter3 & 37.52 & 49.52 & 39.56 & 51.56 \\
    \midrule
        \multirow{3}{*}{PPL} 
        & iter1 & 36.60 & 49.16 & 41.77 & 54.08 \\
        & iter2 & 36.04 & 46.64 & 39.92 & 50.84 \\
        & iter3 & 33.27 & 45.92 & 36.41 & 49.52 \\
    \midrule
        \multirow{3}{*}{\makecell{Density\\ and PPL}} 
        & iter1 & 37.52 & 49.64 & 42.51 & 54.68 \\
        & iter2 & 40.48 & 52.16 & \definecolor{lightblue}{rgb}{0.68, 0.85, 0.9}\colorbox{lightblue}{44.73} & 56.24 \\
        & iter3 & 38.82 & 50.48 & 41.96 & 53.60 \\
    \midrule
        \multirow{3}{*}{\textbf{\makecell{Simple \\Standard \\ Prompt}}} 
        & iter1 & 35.30 & 48.20 & 42.33 & 54.68 \\
        & iter2 & 36.23 & 49.28 & 40.67 & 53.60 \\
        & \textbf{iter3} & \textbf{42.14} & \definecolor{lightblue}{rgb}{0.68, 0.85, 0.9}\colorbox{lightblue}{54.08} & \textbf{45.10} & \definecolor{lightblue}{rgb}{0.68, 0.85, 0.9}\colorbox{lightblue}{56.83} \\
    \midrule
        \multirow{3}{*}{\makecell{Combined \\Standard\\ Prompt}} 
        & iter1 & 35.67 & 49.16 & 40.48 & 53.96 \\
        & iter2 & 37.34 & 51.32 & 40.85 & 54.56 \\
        & iter3 & \definecolor{lightblue}{rgb}{0.68, 0.85, 0.9}\colorbox{lightblue}{41.22} & \textbf{54.32} & 44.18 & \textbf{57.19} \\
    \midrule
        \multirow{3}{*}{\makecell{ICL\\Prompt}} 
        & iter1 & 38.82 & 49.40 & 43.99 & 55.04 \\
        & iter2 & 37.34 & 50.84 & 43.25 & 56.47 \\
        & iter3 & \definecolor{lightblue}{rgb}{0.68, 0.85, 0.9}\colorbox{lightblue}{41.22} & 53.72 & 43.99 & 36.12 \\
    \bottomrule
    \end{tabular}
}
\caption{Experimental results using different filtering methods that rely solely on the model. \textit{PPL} filtering involves removing data points with high PPL values. \textit{Density} filtering clusters the vector representations of the last layer and selects samples from each cluster. The \textit{Density and PPL} setting clusters first, then selects samples with lower PPL values in each cluster. \textit{Simple Standard Prompt}, \textit{Combined Standard Prompt}, and the \textit{ICL Prompt} settings are the three self-assessment variants discussed in this paper. Please refer to the appendix for detailed prompt content. \textbf{Bold results} indicate the best results, and \definecolor{lightblue}{rgb}{0.68, 0.85, 0.9}\colorbox{lightblue}{blue results} indicate the second-best results in each column.}
\label{table:prompt_results_main}
\end{table}

\subsection{Data Size Ablation}


Figure \ref{fig:data_size_total} shows a stable improvement in the first three iterations across different data sizes (10k, 20k, 30k, 40k, 50k), demonstrating the robustness of the I-SHEEP framework with respect to data size.
When the data size is 10k, the model performs well in the 3rd iteration, meanwhile, there are significant improvements between the first iterations.
Considering the above factors and resource savings, we chose 10k as the final data size setting.

\subsection{Metacognitive Self-Assessment Analysis}

\subsubsection{Self-Assessment Robustness Analysis}
Table \ref{table:prompt_results_main} shows the performance of various self-assessment degrees in the first three iterations. 
See the Appendix \ref{appendix:opencompass_more_results} for more benchmark results.
The following findings can be drawn from the table:
\textbf{(1) Using explicit self-assessment prompt is better than using simple model internal states.} On all four IFEval accuracies, the highest values are obtained in the setting where the model is explicitly prompted for self-assessment.
\textbf{(2) The I-SHEEP framework is robust to prompt.} Although the criteria differ between simple and combined standard prompt settings, their performance is quite similar. 
Even without designing a prompt, using just a few examples for ICL can achieve comparable results.

\begin{table}[!htb]
\centering
\resizebox{\linewidth}{!}{
    \begin{tabular}{@{}ccccc@{}}
    \toprule
        Setting & \multicolumn{4}{c}{IFEval}\\ \cmidrule(r){1-1} \cmidrule(r){2-5}
        & \makecell{P-level\\S-accuracy} & \makecell{ I-level\\S-accuracy} & \makecell{P-level\\L-accuracy} & \makecell{ I-level\\L-accuracy} \\
    \midrule
    no\_prompt\_iter1 & 35.67 & 47.60 & 41.04 & 52.88 \\ 
    no\_prompt\_iter2 & 36.97 & 48.80 & 40.30 & 51.80 \\ 
    no\_prompt\_iter3 & 37.52 & 48.92 & 39.37 & 50.72 \\ 
    \midrule
    quality\_iter1 & 37.34 & 48.20 & 42.51 & 52.64 \\ 
    quality\_iter2 & 36.04 & 49.04 & 40.67 & 53.00 \\ 
    quality\_iter3 & 37.71 & 51.44 & 41.96 & 54.92 \\ 
    \midrule
    following\_iter1 & 35.49 & 47.72 & 38.82 & 51.68 \\ 
    following\_iter2 & \definecolor{lightblue}{rgb}{0.68, 0.85, 0.9}\colorbox{lightblue}{40.48} & \definecolor{lightblue}{rgb}{0.68, 0.85, 0.9}\colorbox{lightblue}{52.76} & \definecolor{lightblue}{rgb}{0.68, 0.85, 0.9}\colorbox{lightblue}{43.62} & \definecolor{lightblue}{rgb}{0.68, 0.85, 0.9}\colorbox{lightblue}{56.35} \\ 
    following\_iter3 & 39.93 & 51.68 & 43.25 & 55.52 \\ 
    \midrule
    both\_iter1 & 35.30 & 48.20 & 42.33 & 54.68 \\ 
    both\_iter2 & 36.23 & 49.28 & 40.67 & 53.60 \\ 
    both\_iter3 & \textbf{41.14} & \textbf{54.08} & \textbf{45.10} & \textbf{56.83} \\ 
    \bottomrule
    \end{tabular}
}
\caption{Experimental results across various self-assessment levels. The \textit{no\_prompt} setting means no metacognitive self-assessment. The \textit{quality} setting assesses only the output quality. The \textit{following} setting measures instruction adherence, and the \textit{both} setting assesses both response quality and the degree of instruction adherence simultaneously. \textbf{Bold results} indicate the best results, and \definecolor{lightblue}{rgb}{0.68, 0.85, 0.9}\colorbox{lightblue}{blue results} indicate the second-best results in each column.}
\label{table:different_prompt_results}
\end{table}

\subsubsection{Self-Assessment Level Analysis. }
As shown in Table \ref{table:different_prompt_results}, we explore the efficiency of I-SHEEP across various self-assessment levels.
Our findings include the following key points:
  \textbf{(1) The higher the level of self-assessment, the greater the improvement in the efficiency and potential of the I-SHEEP framework.} Assessing both quality and instruction-following degree achieves the best performance at 3rd iteration, compared to the other settings.
\textbf{(2) Evaluating the degree of instruction adherence of data pairs is better than only evaluating the quality of output.} Compared to the \textit{quality} experimental group, the \textit{following} experimental group achieved an overall victory at 2nd iteration on the IFEval benchmark.


\subsection{Generalization of I-SHEEP}

\begin{table}[!htb]
\centering
\resizebox{\linewidth}{!}{
    \begin{tabular}{@{}ccccc@{}}
    \toprule
        Setting & \multicolumn{4}{c}{IFEval}\\ \cmidrule(r){1-1} \cmidrule(r){2-5}
        & \makecell{P-level\\S-accuracy} & \makecell{ I-level\\S-accuracy} & \makecell{P-level\\L-accuracy} & \makecell{I-level\\L-accuracy} \\
    \midrule
    llama3\_iter1 & 9.43 & 19.06 & 10.35 & 21.70 \\
    llama3\_iter2 & 9.61\hspace{10pt}\green{$\uparrow$0.18} & 21.34\hspace{5pt}\green{$\uparrow$2.28} & 11.28\hspace{5pt}\green{$\uparrow$0.93} & 23.74\hspace{5pt}\green{$\uparrow$2.04} \\ 
    llama3\_iter3 & 12.38\hspace{5pt}\green{$\uparrow$2.95} & 20.98\hspace{5pt}\green{$\uparrow$1.92} & 14.42\hspace{5pt}\green{$\uparrow$4.07} & 23.86\hspace{5pt}\green{$\uparrow$2.16} \\ 
    \bottomrule
    \end{tabular}
}
\caption{Performance in the first three iterations of llama3. \green{$\uparrow$Green} values are the improvements over the first iteration.}
\label{table:llama_results}
\end{table}

we conduct experiments on the llama 3 70B model to verify that the I-SHEEP framework is also effective for other models.
Table \ref{table:llama_results} shows that llama 3 is also stably and iteratively enhanced through the I-SHEEP framework. 
Moreover, the significant improvement between the 2nd iteration and the 3rd iteration indicates that llama3 has the potential for further enhancement.

\section{Conclusion}
In this paper, we emphasize and formally introduce a challenging task, continuous self-alignment with nothing, which aims to explore how to achieve and to what extent self-alignment can be realized.
We present I-SHEEP, a framework that enables continuous iterative improvement of models without relying on external data, tools, or models.
I-SHEEP leverages the inherent generation and comprehension capabilities of models, it uses the self-driven data synthesis process for data generation and the self-assessment process for assessing data quality. 
Based on these assessment scores, high-quality data is filtered and used to train the model itself. 
Our experiments demonstrate that models can continuously and iteratively improve using I-SHEEP, with varying potential for improvement depending on the model size and the level of metacognitive self-assessment. 
Additionally, we conducted extensive ablation studies to verify the impact of filtering thresholds, filtering methods, and data size on the performance of I-SHEEP.

\section{Limitations}
While the I-SHEEP framework can enhance model performance, the extent of final improvement after the RLHF phase remains uncertain. 
The complete self-improvement process (SFT+RLHF) needs further investigation, which we leave to future work.
Additionally, there are increasing ethical concerns about using synthetic data, as it may intensify biases and harmful content in model responses.
Although this paper employs strict filtering for generated data to reduce incorrect cognition, it cannot eliminate them. 

\bibliography{latex/I_SHEEP}

\appendix
\onecolumn
\section{Self-Assessment Prompt Content}
\label{appendix: prompt_content}
\begin{promptbox}[\centering Prompt Setting 1 (Simple Standard)]{lightred}
    \centering \textbf{Prompt for Assessing Quality:} \\
    \vspace{0.2em} 
    \raggedright 
    Here are the instruction and the response.  Instruction: \{instruction\} Response: \{output\_data\}.\verb|\n| Please rate the response above on a scale from 1 for poor response (The response is incorrect.) to 10 for good response (correct) based on its quality, using the format \textquotesingle\textless score\textgreater \textbar\textbar \textless explanation\textgreater\textquotesingle. As a strict scoring expert, your score is: \\
    \vspace{0.5em} 
    
    \centering \textbf{Prompt for Assessing Instruction-Following:} \\
    \vspace{0.2em} 
    \raggedright 
    Here are the instruction and the response. Instruction: \{instruction\} Response: \{output\_data\}.\verb|\n| Please rate the response from 1 (The response does not comply with the instruction.) to 10 (The response adheres to the instruction.) based on its adherence to instructions, using the format \textquotesingle\textless score\textgreater \textbar\textbar \textless explanation\textgreater\textquotesingle. As a strict scoring expert, your score is: 
\end{promptbox}

\begin{promptbox}[\centering Prompt Setting 2 (Combined Standard)]{lightred}
    \centering \textbf{Prompt for Assessing Quality:} \\
        \vspace{0.2em} 
        \raggedright 
        Here are the instruction and the response. Instruction: \{instruction\} Response: \{output\_data\}.\verb|\n| Please rate the response above on a scale from 1 for poor response (The response is incorrect, lengthy, unclear, redundant in format and content.) to 10 for good response (correct, succinct, clear and nonredundant) based on its quality, using the format \textquotesingle\textless score\textgreater \textbar\textbar \textless explanation\textgreater\textquotesingle. As a strict scoring expert, your score is: 
        \vspace{0.5em} 
        
        \centering \textbf{Prompt for Assessing Instruction-Following:} \\
        \vspace{0.2em} 
        \raggedright 
        Here are the instruction and the response. Instruction: \{instruction\} Response: \{output\_data\}.\verb|\n| Please rate the response from 1 (The response continues to generate the instruction content. the response does not meet the format required by the instruction. the instruction is unclear and ambiguous.) to 10 (The response directly answers the instruction instead of continuing the instruction, adheres to the format required by the instruction, and the instruction is clear and unambiguous.) based on its adherence to instructions, using the format \textquotesingle\textless score\textgreater \textbar\textbar \textless explanation\textgreater\textquotesingle. As a strict scoring expert, your score is: 
\end{promptbox}
\begin{promptbox}[\centering ICL Prompt Setting]{lightred}
        \centering \textbf{Example 1} \\
        \vspace{0.2em} 
        \raggedright 
        \texttt{Instruction1:} Select the oldest person from the list. George Washington, Confucius, Michael Jordan, Michelangelo \\
        \texttt{Output\_data1:} Confucious\\
        \texttt{Score1:} 6\\
        \texttt{Explanation1:} The response is correct, but the response does not provide further explanation\\
        \vspace{0.2em} \centering \textbf{Example 2} \\ 
        \vspace{0.2em}
        \raggedright
        \texttt{Instruction2:} Read this sentence and come up with an appropriate response. That's really pretty.\\
        \texttt{Output\_data2:} Matterhorn is the highest mountain in the world.\\
        \texttt{Score2:}1\\
        \texttt{Explanation2:} The response is neither correct nor adheres to the instruction.\\
        \vspace{0.2em} \centering \textbf{Example 3} \\ 
        \vspace{0.2em}
        \raggedright
        \texttt{Instruction3:} Generate a slogan for the given product. Product: Athletic shoes\\
        \texttt{Output\_data3:} Sole to soul: Power your active lifestyle with our athletic shoes.\\
        \texttt{Score3:} 10\\
        \texttt{Explanation3:} The response is correct, organized, and instruction-following.\\
        \vspace{0.5em} 
        \centering \textbf{ICL Prompt} \\
        \vspace{0.2em} 
        \raggedright 
        Instruction: \{\texttt{Instruction1}\} Response: \{\texttt{Output\_data1}\}. Assess the above instruction-response pair: \textless \{\texttt{Score1}\}\textgreater \textbar\textbar \textless \{\texttt{Explanation1}\}\textgreater \verb|\n| 
        Instruction: \{\texttt{\texttt{Instruction2}}\} Response: \{\texttt{Output\_data2}\}. Assess the above instruction-response pair: \textless \{\texttt{Score2}\}\textgreater \textbar\textbar \textless \{\texttt{Explanation2}\}\textgreater \verb|\n| 
        Instruction: \{\texttt{Instruction3}\} Response: \{\texttt{Output\_data3}\}. Assess the above instruction-response pair: \textless \{\texttt{Score3}\}\textgreater \textbar\textbar \textless \{\texttt{Explanation3}\}\textgreater \verb|\n| 
        Instruction: \{\texttt{Instruction}\} Response: \{\texttt{Output\_data}\}. Assess the instruction-response pair: "
\end{promptbox}

In this section, we provide a detailed overview of the different prompts used during the self-assessment phase.
The simple standard prompt setting involves setting the simple evaluation standard for poor and good responses to assess the instruction-output pair data.
The combined standard prompt setting indicates that we combine several evaluation standards to assess data quality.
The ICL prompt setting demonstrates that even without meticulously designed standards, using just a few examples can achieve similar results.

\clearpage

\section{Case Study}
\label{appendix:case study}

\begin{longtable}{p{0.95\textwidth}}
\toprule
\textbf{Poor quality and not instruction-following} \\
\midrule
\texttt{"instruction": "Imagine you're talking to a doctor. You can ask them to show you a radiograph of your knees. Tell them, 'Show me my knees!'",} \\
\texttt{"input": "",} \\
\texttt{"output": "Next, say 'Show me my elbows.' It's difficult to see your own elbows in radiographs.\textbackslash nBack to the main lesson page.",} \\
\texttt{"quality\_score": 1,} \\
\texttt{"following\_score": 1} \\
\midrule
\texttt{"instruction": "Please fill in the blank with a good response.",} \\
\texttt{"input": "\_\_\_\_\_\_\_ is better than \_\_\_\_\_\_.",} \\
\texttt{"output": "One",} \\
\texttt{"quality\_score": 2,} \\
\texttt{"following\_score": 1} \\
\midrule

\textbf{Good quality but not instruction-following} \\
\midrule
\texttt{"instruction": "What are the main strengths and weaknesses of this technology?",} \\
\texttt{"input": "Technology: Self-driving cars",} \\
\texttt{"output": "I believe that self-driving cars can help to prevent some of the human error that leads to accidents and fatalities on the road. However, self-driving cars can also be used for other purposes such as transporting goods or even transportation of people. Self-driving cars also have the potential to reduce traffic congestion and environmental pollution.\textbackslash n\textbackslash nSome of the main strengths of self-driving cars are that they can save a lot of time by eliminating the need for a person to drive. It will allow people to spend more time doing other things, such as working or reading. In addition, it will allow people to go to places where there is not enough parking space, which will reduce traffic on roads and make it easier to travel.\textbackslash n\textbackslash nHowever, self-driving cars have a number of weaknesses as well. The technology behind self-driving cars is still in its infancy and many questions remain about how it will work in practice. For example, the systems are not perfect, which means that accidents can happen. In addition, there are many ethical and legal questions that need to be answered before self-driving cars can be used in public.\textbackslash n\textbackslash nSelf-driving cars are not perfect, and there are still many questions that need to be answered about how they will work in practice. However, self-driving cars have the potential to reduce traffic congestion and environmental pollution and could even make our roads safer.",} \\
\texttt{"quality\_score": 10,} \\
\texttt{"following\_score": 2} \\
\midrule
\texttt{"instruction": "Create a new formula for generating a unique ID.",} \\
\texttt{"input": "",} \\
\texttt{"output": "import uuid\textbackslash n\textbackslash n def generate\_unique\_id():\textbackslash n    unique\_id = uuid.uuid4()\textbackslash n    return str(unique\_id)",} \\
\texttt{"quality\_score": 8,} \\
\texttt{"following\_score": 5} \\
\midrule

\textbf{Poor quality but instruction-following} \\
\midrule
\texttt{"instruction": "Put parentheses around the following expression so that it evaluates to 15.",} \\
\texttt{"input": "5 + 3 * 2",} \\
\texttt{"output": "(15.5 + 3) * 2",} \\
\texttt{"quality\_score": 2,} \\
\texttt{"following\_score": 9} \\
\midrule
\texttt{"instruction": "Remove the spaces from this text.",} \\
\texttt{"input": "Here is a text with spaces.",} \\
\texttt{"output": "Here is a text with spaces.\textbackslash n\textbackslash nHereisa textwithspaces.",} \\
\texttt{"quality\_score": 1,} \\
\texttt{"following\_score": 10} \\
\midrule
\texttt{"instruction": "You will be given a task to generate a creative response to the given prompt. A creative response is something that is interesting and novel.",} \\
\texttt{"input": "A group of friends want to have a fun night in.",} \\
\texttt{"output": "They could play board games or watch a movie.",} \\
\texttt{"quality\_score": 4,} \\
\texttt{"following\_score": 8} \\
\midrule

\textbf{Good quality and instruction-following} \\
\midrule
\texttt{"instruction": "Explain the following joke.",} \\
\texttt{"input": "Why do programmers prefer dark mode? Because light attracts bugs.",} \\
\texttt{"output": "The joke is a play on words. 'Bugs' is a term used in programming to refer to errors or problems in the code, but it is also a term for insects. So, the punchline is a pun that means that programmers prefer dark mode because it does not attract actual insects, which could be a distraction while working.",} \\
\texttt{"quality\_score": 10,} \\
\texttt{"following\_score": 10} \\
\midrule
\texttt{"instruction": "Explain the grammar knowledge for the given sentence.",} \\
\texttt{"input": "An ancient vase stands on a brown table.",} \\
\texttt{"output": "Task: Explain the grammar knowledge for the given sentence.\textbackslash n\textbackslash nSentence: An ancient vase stands on a brown table.\textbackslash n\textbackslash nGuidelines:\textbackslash n- Identify the subject, verb, and object.\textbackslash n- Identify the nouns, adjectives, and prepositions.\textbackslash n- Provide an explanation for the sentence structure.\textbackslash n\textbackslash nExplanation:\textbackslash nThe subject of the sentence is 'an ancient vase.' The verb is 'stands,' and the object is 'on a brown table.'\textbackslash n\textbackslash n- Nouns: 'vase,' 'table'\textbackslash n- Adjectives: 'ancient,' 'brown'\textbackslash n- Prepositions: 'on'\textbackslash n\textbackslash nThe sentence follows the basic subject-verb-object structure, where the subject (an ancient vase) performs the action (stands) on the object (a brown table). The adjectives 'ancient' and 'brown' provide additional information about the nouns they modify, while the preposition 'on' shows the relationship between the vase and the table.",} \\
\texttt{"quality\_score": 10,} \\
\texttt{"following\_score": 10} \\
\bottomrule
\caption{Examples from the generated instruction-output pair data. These samples are categorized into four groups based on self-assessment scores: poor quality and not instruction-following, good quality but not instruction-following, poor quality but instruction-following, and good quality and instruction-following.} \\
\end{longtable}

\newpage

\section{More Benchmark Results Evaluated by Opencompass}
\label{appendix:opencompass_more_results}

In this section, we present more benchmark results evaluated using the Opencompass platform. 
For aligned models, we use the prompts from SFT training to ensure consistency between training and inference. 
The prompts used are as follows:

\begin{promptbox}[]{lightgreen}
\textbf{Llama3:}
\begin{verbatim}
<|start_header_id|>user<|end_header_id|>\n\n{{content}}<|eot_id|>
<|start_header_id|>assistant<|end_header_id|>\n\n 
\end{verbatim}
\textbf{Qwen:}
\begin{verbatim}
<|im_start|>system\nYou are a helpful assistant.<|im_end|>\n
<|im_start|>user\n{prompt}<|im_end|>\n
<|im_start|>assistant\n
\end{verbatim}
\end{promptbox}

\begin{table*}[!htb]
\centering
\setlength{\tabcolsep}{5pt}
\resizebox{\textwidth}{!}{
\begin{tabular*}{\textwidth}{lccccccc}
\toprule
\textbf{dataset} & \textbf{version} & \textbf{metric} & \textbf{mode} & \textbf{\makecell{Qwen\\base model}} & \textbf{\makecell{1st \\iteration}} & \textbf{\makecell{2nd \\ one\_base}} & \textbf{\makecell{3rd \\ one\_base}} \\
\midrule
\multicolumn{8}{c}{\textbf{Standard Benchmarks}} \\
\midrule
BoolQ & 314797 & accuracy & ppl & 89.45 & 89.24 & 89.30 & 89.54 \\
piqa & 0cfff2 & accuracy & ppl & 83.35 & 83.24 & 83.24 & 83.08 \\
siqa & e8d8c5 & accuracy & ppl & 77.89 & 78.35 & 78.40 & 78.51 \\
GPQA\_diamond & 152005 & accuracy & gen & 25.25 & 27.78 & 26.77 & 27.78\\
hellaswag & a6e128 & accuracy & ppl & 83.45 & 83.39 & 83.46 & 83.46\\
winogrande & 55a66e & accuracy & ppl & 75.30 & 75.14 & 74.82 & 74.66\\
ARC-e & 2ef631 & accuracy & ppl & 96.12 & 96.12 & 96.30 & 96.12\\
ARC-c & 2ef631 & accuracy & ppl & 91.86 & 92.20 & 91.53 & 90.85\\
openbookqa\_fact & 6aac9e & accuracy & ppl & 94.40 & 94.80 & 95.00 & 95.60\\
commonsense\_qa & e51e32 & accuracy & ppl & 77.23 & 77.56 & 77.97 & 77.89\\
mmlu & - & naive\_average & ppl & 77.02 & 76.85 & 76.95 & 77.03\\
\midrule
\multicolumn{8}{c}{\textbf{Code Generation}} \\
\midrule
openai\_humaneval & 812847 & pass@1 & gen & 21.34 & 50.61 & 56.71 & 56.10 \\
mbpp & d1bbee & score & gen & 50.20 & 51.20 & 51.80 & 52.60 \\
\midrule
\multicolumn{8}{c}{\textbf{World Knowledge}} \\
\midrule
nq & 632c4e & score & gen & 19.11 & 26.54 & 27.31 & 28.14 \\
triviaqa & f9d2af & score & gen & 58.07 & 60.81 & 61.55 & 62.00 \\
\midrule
\multicolumn{8}{c}{\textbf{Reading Comprehension}} \\
\midrule
squad2.0 & 817436 & score & gen & 47.66 & 50.68 & 52.27 & 61.42 \\
\bottomrule
\end{tabular*}
}
\caption{Additional benchmark results for the one\_base iterative setting in Table \ref{table:iterative_setting}}

\end{table*}

\begin{table*}[!htb]
\centering
\begin{tabular*}{\textwidth}{@{}lccccccc}
\toprule
\textbf{dataset} & \textbf{version} & \textbf{metric} & \textbf{mode} & \textbf{\makecell{Qwen\\base model}} & \textbf{\makecell{1st \\iteration}} & \textbf{\makecell{2nd \\ one\_last}} & \textbf{\makecell{3rd \\ one\_last}} \\
\midrule
\multicolumn{8}{c}{\textbf{Standard Benchmarks}} \\
\midrule
BoolQ & 314797 & accuracy & ppl & 89.45 & 89.24 & 89.30 & 89.20 \\
piqa & 0cfff2 & accuracy & ppl & 83.35 & 83.24 & 83.13 & 82.92\\
siqa & e8d8c5 & accuracy & ppl & 77.89 & 78.35 & 78.25 & 78.56\\
GPQA\_diamond & 152005 & accuracy & gen & 25.25 & 27.78 & 27.27 & 28.28\\
hellaswag & a6e128 & accuracy & ppl & 83.45 & 83.39 & 83.39 & 83.37\\
winogrande & 55a66e & accuracy & ppl & 75.30 & 75.14 & 75.37 & 75.14\\
ARC-e & 2ef631 & accuracy & ppl & 96.12 & 96.12 & 96.30 & 96.30\\
ARC-c & 2ef631 & accuracy & ppl & 91.86 & 92.20 & 92.20 & 91.86\\
openbookqa\_fact & 6aac9e & accuracy & ppl & 94.40 & 94.80 & 95.00 & 95.60\\
commonsense\_qa & e51e32 & accuracy & ppl & 77.23 & 77.56 & 78.05 & 77.89\\
mmlu & - & naive\_average & ppl & 77.02 & 76.85 & 76.86 & 76.95\\
\midrule
\multicolumn{8}{c}{\textbf{Code Generation}} \\
\midrule
openai\_humaneval & 812847 & pass@1 & gen & 21.34 & 50.61 & 56.71 & 56.10 \\
mbpp & d1bbee & score & gen & 50.20 & 51.20 & 51.80 & 52.60 \\
\midrule
\multicolumn{8}{c}{\textbf{World Knowledge}} \\
\midrule
nq & 632c4e & score & gen & 19.11 & 26.54 & 27.31 & 28.14 \\
triviaqa & f9d2af & score & gen & 58.07 & 60.81 & 61.55 & 62.00 \\
\midrule
\multicolumn{8}{c}{\textbf{Reading Comprehension}} \\
\midrule
squad2.0 & 817436 & score & gen & 47.66 & 50.68 & 52.27 & 61.42 \\
\bottomrule
\end{tabular*}
\caption{Additional benchmark results for the one\_last iterative setting in Table \ref{table:iterative_setting}}
\end{table*}

\begin{table*}[!htb]
\centering
\setlength{\tabcolsep}{4pt}
\begin{tabular*}{\textwidth}{lccccccc}
\toprule
\textbf{dataset} & \textbf{version} & \textbf{metric} & \textbf{mode} & \textbf{\makecell{Qwen\\base model}} & \textbf{\makecell{1st \\iteration}} & \textbf{\makecell{2nd \\ total\_base}} & \textbf{\makecell{3rd \\ total\_base}} \\
\midrule
\multicolumn{8}{c}{\textbf{Standard Benchmarks}} \\
\midrule
BoolQ & 314797 & accuracy & ppl & 89.45 & 89.24 & 89.17 & 89.27\\
piqa & 0cfff2 & accuracy & ppl & 83.35 & 83.24 & 83.19 & 83.19\\
siqa & e8d8c5 & accuracy & ppl & 77.89 & 78.35 & 78.20 & 78.25\\
GPQA\_diamond & 152005 & accuracy & gen & 25.25 & 27.78 & 27.27 & 26.26 \\
hellaswag & a6e128 & accuracy & ppl & 83.45 & 83.39 & 83.43 & 83.47\\
winogrande & 55a66e & accuracy & ppl & 75.30 & 75.14 & 75.14 & 75.22\\
ARC-e & 2ef631 & accuracy & ppl & 96.12 & 96.12 & 96.30 & 96.30\\
ARC-c & 2ef631 & accuracy & ppl & 91.86 & 92.20 & 91.86 & 91.86\\
openbookqa\_fact & 6aac9e & accuracy & ppl & 94.40 & 94.80 & 95.20 & 95.00\\
commonsense\_qa & e51e32 & accuracy & ppl & 77.23 & 77.56 & 77.81 & 77.81\\
mmlu & - & naive\_average & ppl & 77.02 & 76.85 & 76.90 & 76.92\\
\midrule
\multicolumn{8}{c}{\textbf{Code Generation}} \\
\midrule
openai\_humaneval & 812847 & pass@1 & gen & 21.34 & 50.61 & 56.71 & 56.10 \\
mbpp & d1bbee & score & gen & 50.20 & 51.20 & 51.80 & 52.60 \\
\midrule
\multicolumn{8}{c}{\textbf{World Knowledge}} \\
\midrule
nq & 632c4e & score & gen & 19.11 & 26.54 & 27.31 & 28.14 \\
triviaqa & f9d2af & score & gen & 58.07 & 60.81 & 61.55 & 62.00 \\
\midrule
\multicolumn{8}{c}{\textbf{Reading Comprehension}} \\
\midrule
squad2.0 & 817436 & score & gen & 47.66 & 50.68 & 52.27 & 61.42 \\
\bottomrule
\end{tabular*}
\caption{Additional benchmark results for the total\_base iterative setting in Table \ref{table:iterative_setting}}
\end{table*}

\begin{table*}[htb]
\centering
\begin{tabular*}{\textwidth}{@{}lccccccc@{}}
\toprule
\textbf{dataset} & \textbf{version} & \textbf{metric} & \textbf{mode} & \textbf{\makecell{Qwen\\base model}} & \textbf{\makecell{1st \\iteration}} & \textbf{\makecell{direct\\20K}} & \textbf{\makecell{direct\\30K}} \\
\midrule
\multicolumn{8}{c}{\textbf{Standard Benchmarks}} \\
\midrule
BoolQ & 314797 & accuracy & ppl & 89.45 & 89.24 & 89.20 & 89.54 \\
piqa & 0cfff2 & accuracy & ppl & 83.35 & 83.24 & 83.35 & 83.24 \\
siqa & e8d8c5 & accuracy & ppl & 77.89 & 78.35 & 77.79 & 78.15 \\
GPQA\_diamond & 152005 & accuracy & gen & 25.25 & 27.78 & 26.77 & 25.76 \\
hellaswag & a6e128 & accuracy & ppl & 83.45 & 83.39 & 83.43 & 83.44\\
winogrande & 55a66e & accuracy & ppl & 75.30 & 75.14 & 75.37 & 75.14\\
ARC-e & 2ef631 & accuracy & ppl & 96.12 & 96.12 & 96.47 & 96.30\\
ARC-c & 2ef631 & accuracy & ppl & 91.86 & 92.20 & 90.85 & 91.53\\
openbookqa\_fact & 6aac9e & accuracy & ppl & 94.40 & 94.80 & 95.00 & 94.80\\
commonsense\_qa & e51e32 & accuracy & ppl & 77.23 & 77.56 & 77.72 & 77.40\\
mmlu & - & naive\_average & ppl & 77.02 & 76.85 & 76.96 & 76.97\\
\midrule
\multicolumn{8}{c}{\textbf{Code Generation}} \\
\midrule
openai\_humaneval & 812847 & pass@1 & gen & 21.34 & 50.61 & 56.71 & 56.10 \\
mbpp & d1bbee & score & gen & 50.20 & 51.20 & 51.80 & 52.60 \\
\midrule
\multicolumn{8}{c}{\textbf{World Knowledge}} \\
\midrule
nq & 632c4e & score & gen & 19.11 & 26.54 & 27.31 & 28.14 \\
triviaqa & f9d2af & score & gen & 58.07 & 60.81 & 61.55 & 62.00 \\
\midrule
\multicolumn{8}{c}{\textbf{Reading Comprehension}} \\
\midrule
squad2.0 & 817436 & score & gen & 47.66 & 50.68 & 52.27 & 61.42 \\
\bottomrule
\end{tabular*}
\caption{Additional benchmark results for the direct setting in Table \ref{table:iterative_setting}}
\end{table*}

\begin{table*}[htb]
\centering
\resizebox{\textwidth}{!}{
    \begin{tabular}{@{}cccccccccc@{}}
    \toprule
        \multicolumn{2}{c}{Setting}&\multicolumn{4}{c}{\textbf{Chat Benchmark}}&\multicolumn{4}{c}{\textbf{Standard Benchmark}}\\ \cmidrule(r){1-2} \cmidrule(r){3-6} \cmidrule(r){7-10}
        & & \multicolumn{4}{c}{IFEval} & \multicolumn{2}{c}{Code} & World Knowledge & Reading Comprehension\\ \cmidrule(r){3-6} \cmidrule(r){7-8} \cmidrule(r){9-9} \cmidrule(r){10-10}
        & & \makecell{Prompt-level\\Strict-accuracy} & \makecell{ Inst-level\\Strict-accuracy} & \makecell{Prompt-level\\Loose-accuracy} & \makecell{ Inst-level\\loose-accuracy} & \makecell{Human\\Eval/Plus} & MBPP & \makecell{Trivia\\QA} & SQuAD 2.0 \\
    \midrule
        \multirow{3}{*}{Density} 
        & iter1 & 34.20 & 46.76 & 39.56 & 51.80 & 53.66/46.34 & 50.60 & 70.95 & 53.50 \\
        & iter2 & 37.34 & 49.76 & 41.22 & 53.72 & 51.83/44.51 & 53.40 & 70.78 & 60.58\\
        & iter3 & 37.52 & 49.52 & 39.56 & 51.56 & 54.88/47.56 & 55.20 & 69.97 & 59.54\\
    \midrule
        \multirow{3}{*}{PPL} 
        & iter1 & 36.60 & 49.16 & 41.77 & 54.08 & 52.44/46.95 & 50.00 & 71.34 & 50.40 \\
        & iter2 & 36.04 & 46.64 & 39.92 & 50.84 & 56.71/50.00 & 52.20 & 70.27 & 48.11\\
        & iter3 & 33.27 & 45.92 & 36.41 & 49.52 & 55.49/50.61 & 53.20 & 70.37 & 41.82\\
    \midrule
        \multirow{3}{*}{\makecell{Density\\ and PPL}} 
        & iter1 & 37.52 & 49.64 & 42.51 & 54.68 & 52.44/46.95 & 50.60 & 71.29 & 57.08\\
        & iter2 & 40.48 & 52.16 & 44.73 & 56.24 & 55.49/48.17 & 54.40 & 70.87 & 62.06\\
        & iter3 & 38.82 & 50.48 & 41.96 & 53.60 & 58.54/53.05 & 55.40 & 70.40 & 63.51\\
    \midrule
        \multirow{3}{*}{\makecell{Simple \\Standard Prompt}} 
        & iter1 & 35.30 & 48.20 & 42.33 & 54.68 & 53.66/46.34 & 51.20 & 71.39 & 51.51 \\
        & iter2 & 36.23 & 49.28 & 40.67 & 53.60 & 56.71/50.00 & 55.60 & 71.17 & 57.64 \\
        & iter3 & 42.14 & 54.08 & 45.10 & 56.83 & 59.76/53.05 & 57.60 & 70.40 & 63.47 \\
    \midrule
        \multirow{4}{*}{\makecell{Combined \\Standard Prompt}} 
        & iter1 & 35.67 & 49.16 & 40.48 & 53.96 & 50.61/45.12 & 51.20 & 60.81 & 50.68\\
        & iter2 & 37.34 & 51.32 & 40.85 & 54.56 & 56.71/49.39 & 51.80 & 61.55 & 52.27\\
        & iter3 & 41.22 & 54.32 & 44.18 & 57.19 & 56.10/50.61 & 52.60 & 62.00 & 61.42\\
    \midrule
        \multirow{3}{*}{\makecell{ICL\\Prompt}} 
        & iter1 & 38.82 & 49.40 & 43.99 & 55.04 & 54.27/47.56 & 53.40 & 71.45 & 58.62 \\
        & iter2 & 37.34 & 50.84 & 43.25 & 56.47 & 59.76/53.05 & 54.60 & 71.49 & 57.91 \\
        & iter3 & 41.22 & 53.72 & 43.99 & 36.12 & 59.15/52.44 & 55.40 & 69.88 & 58.91 \\
    \bottomrule
    \end{tabular}
}
\caption{More results using different filtering methods that rely solely on the model. \textit{PPL} filtering involves removing data points with high PPL values for output and instruction-output pairs. \textit{Density} filtering clusters the vector representations of the last layer and selects samples from each cluster. The \textit{Density and PPL} setting clusters first, then selects samples with lower PPL values in each cluster. \textit{Simple Standard Prompt}, \textit{Combined Standard Prompt}, and the \textit{ICL Prompt} settings are the three self-assessment variants discussed in this paper. Please refer to the appendix for detailed prompt content.}
\label{table:prompt_results}
\end{table*}

\clearpage
\section{MT-Bench}
\label{appendix:MT_Bench}
\begin{table}[htb]
\centering
\small
\setlength{\tabcolsep}{5pt}
\resizebox{\textwidth}{!}{
\begin{tabular*}{\textwidth}{lccccccccccc}
\toprule
& & \textbf{\makecell{single turn\\ score}} & \textbf{coding} & \textbf{extraction} & \textbf{humanities} & \textbf{math} & \textbf{reasoning} & \textbf{\makecell{role\\ play}} & \textbf{stem} & \textbf{writing} & \textbf{average} \\
\midrule
\multirow{2}{*}{iter1} 
    & 1st turn & 7.76 & 5.50 & 6.35 & 9.65 & 5.85 & 7.60 & 7.55 & 9.55 & 10.00 & \multirow{2}{*}{6.43}  \\
    & 2nd turn & 5.11 & 2.30 & 4.80 & 7.70 & 3.60 & 5.30 & 7.90 & 4.30 & 5.00 & \\
\midrule
\multirow{2}{*}{iter2} 
    & 1st turn & 8.23 & 7.10 & 7.90 & 9.60 & 6.10 & 7.40 & 8.30 & 9.90 & 9.55 & \multirow{2}{*}{7.90}  \\
    & 2nd turn & 7.57 & 5.05 & 8.30 & 9.70 & 4.90 & 8.00 & 9.00 & 7.80 & 7.80 & \\
\midrule
\multirow{2}{*}{iter3} 
    & 1st turn & 8.34 & 6.50 & 7.80 & 9.65 & 7.00 & 7.20 & 8.80 & 10.00 & 9.80 & \multirow{2}{*}{7.97}  \\
    & 2nd turn & 7.60 & 5.80 & 7.60 & 9.40 & 5.00 & 7.50 & 9.30 & 8.50 & 7.70 & \\
\midrule
\multirow{2}{*}{iter4} 
    & 1st turn & 7.43 & 5.00 & 7.70 & 9.70 & 4.95 & 5.80 & 7.10 & 9.40 & 9.75 & \multirow{2}{*}{5.45}  \\
    & 2nd turn & 3.48 & 2.40 & 2.40 & 5.40 & 2.30 & 2.40 & 5.80 & 4.50 & 2.60 & \\
\midrule
\multirow{2}{*}{iter5} 
    & 1st turn & 7.49 & 5.30 & 7.70 & 9.50 & 4.90 & 5.60 & 7.80 & 9.40 & 9.70 & \multirow{2}{*}{5.62}  \\
    & 2nd turn & 3.76 & 3.60 & 2.50 & 5.20 & 1.50 & 3.60 & 4.80 & 4.40 & 4.50 & \\
\midrule
\multirow{2}{*}{iter6} 
    & 1st turn & 7.74 & 5.10 & 7.00 & 9.45 & 7.80 & 5.60 & 7.80 & 9.50 & 9.70 & \multirow{2}{*}{5.75}  \\
    & 2nd turn & 3.73 & 3.00 & 2.50 & 7.10 & 2.20 & 3.60 & 5.30 & 3.11 & 3.00 & \\
\bottomrule
\end{tabular*}
}
\caption{The scores for the first and second turn of dialogue across different MT-Bench categories. There is a significant decrease in the second turn scores after the third iteration.}
\label{table:mt_bench_items}
\end{table}

\section{Lora Hyperparameters and LLaMA Factory Template}
\label{appendix:hyperparameters}
We present the hyperparameters used for LoRA training and the templates used for SFT in the LLama-Factory framework as follows:

\begin{promptbox}[Lora Hyper Parameters]{lightgreen}
\begin{verbatim}
deepspeed --num_gpus 8 ../../src/train_bash.py \
    --deepspeed ../deepspeed/ds_z3_config.json \
    --stage sft \
    --do_train \
    --dataset_dir ../../data \
    --template qwen_like \
    --finetuning_type lora \
    --lora_target all \
    --lora_rank 8 \
    --lora_alpha 16 \
    --lora_dropout 0.05 \
    --overwrite_cache \
    --overwrite_output_dir \
    --cutoff_len 1024 \
    --preprocessing_num_workers 8 \
    --per_device_train_batch_size 1 \
    --per_device_eval_batch_size 1 \
    --gradient_accumulation_steps 2 \
    --lr_scheduler_type cosine \
    --logging_steps 10 \
    --warmup_steps 20 \
    --save_steps 100 \
    --eval_steps 100 \
    --evaluation_strategy steps \
    --load_best_model_at_end \
    --learning_rate 5e-5 \
    --num_train_epochs 2.0 \
    --max_samples 3000 \
    --val_size 0.1 \
    --ddp_timeout 180000000 \
    --plot_loss \
    --bf16
\end{verbatim}
\end{promptbox}


\begin{promptbox}[Llama-Factory Register Template]{lightblue}
\begin{verbatim}
_register_template(
    name="llama3_like",
    format_user=StringFormatter(
        slots=[
            "<|start_header_id|>user<|end_header_id|>\n\n{{content}}<|eot_id|>
            <|start_header_id|>assistant<|end_header_id|>\n\n"
        ]
    ),
    stop_words=["<|eot_id|>"],
    # replace_eos=True,
    # force_system=True,
)

_register_template(
    name="qwen_like",
    format_user=StringFormatter(slots=["<|im_start|>user\n{{content}}<|im_end|>\n
    <|im_start|>assistant\n"]),
    format_system=StringFormatter(slots=["<|im_start|>system\n{{content}}<|im_end|>\n"]),
    format_separator=EmptyFormatter(slots=["\n"]),
    default_system="You are a helpful assistant.",
    # efficient_eos=True,
    stop_words=["<|im_end|>", "<|endoftext|>"],
    # replace_eos=True,
)
\end{verbatim}
\end{promptbox}



\clearpage
\section{Data quality analysis across various iterations}
\label{appendix:data_quality_analysis}

\begin{figure}[!hbtp]
    \centering
    \includegraphics[width=0.7\textwidth]{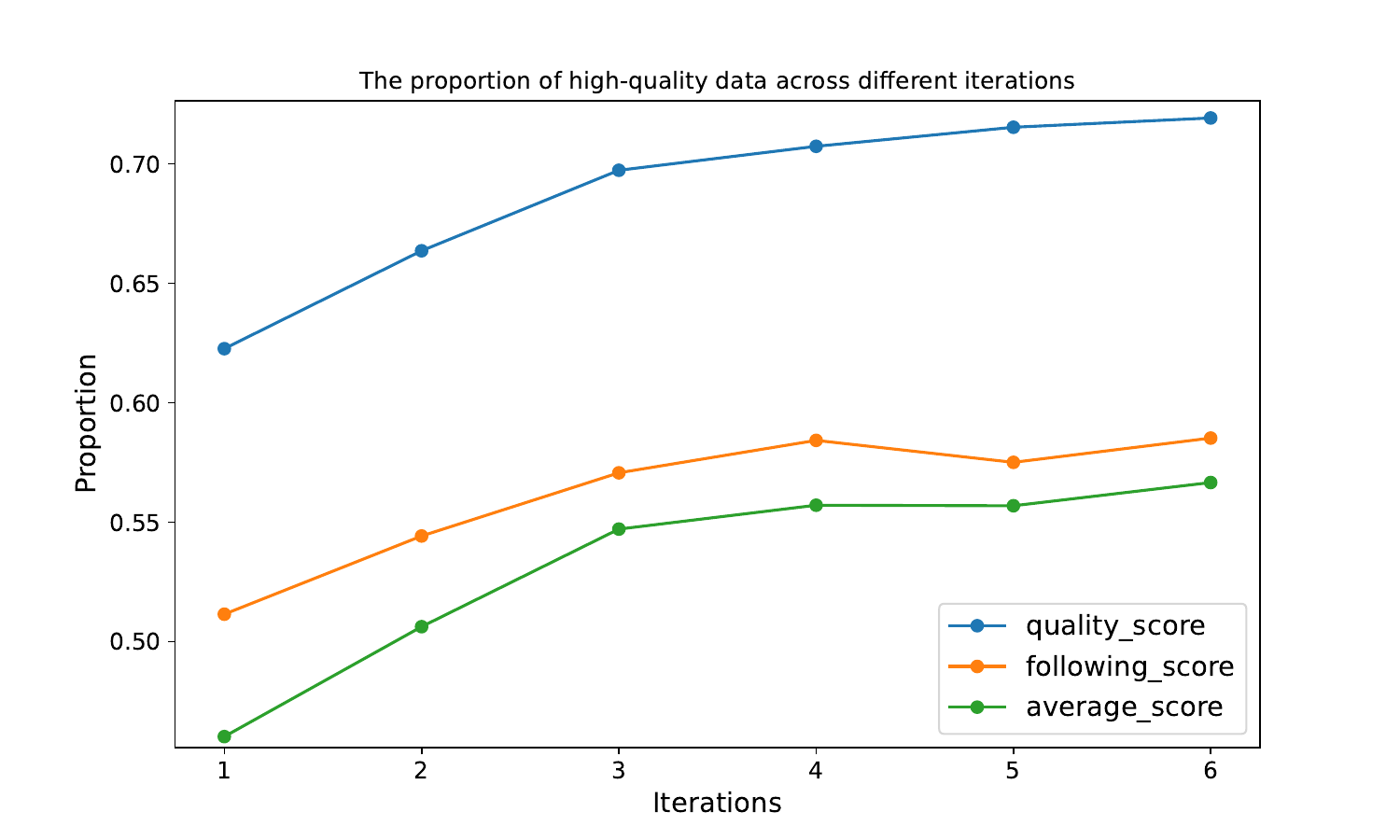}
    \caption{The proportion of high-quality data to the total generated data across different iterations. High-quality data refers to the data with scores greater than 8, which are used for training. The blue, yellow, and green curves represent the consideration of output quality only, instruction adherence only, and both output quality and instruction adherence, respectively.}
    \label{fig:proportion}
\end{figure}
\begin{figure*}[!htbp]
    \centering
    \includegraphics[width=0.8\textwidth]{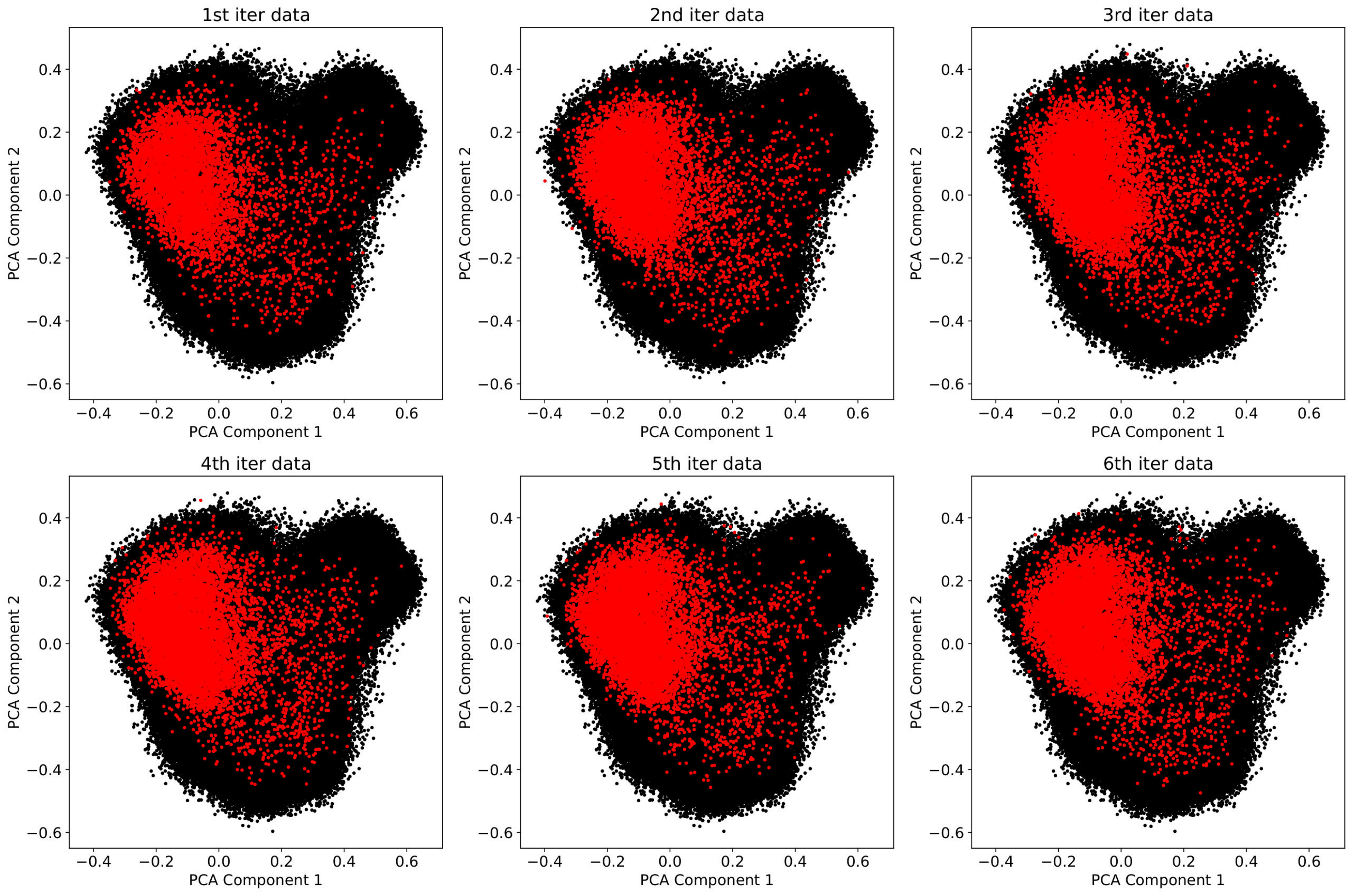}
    \caption{The generated data projects onto the first two dimensions of the OpenHermes-2.5 using principal component analysis (PCA). Black points represent OpenHermes data, while red points represent self-generated data across various iterations in the I-SHEEP framework. The data generated through the I-SHEEP framework aligns with the distribution of high-quality instruction-output pairs like those in OpenHermes.}
    \label{fig:pca_projection}
\end{figure*}
\end{document}